\documentclass{article}
\usepackage[T1]{fontenc}
\usepackage{spconf,amsmath,amssymb,graphicx,hyperref,booktabs,xcolor}

\newcommand{\Mzero}{M_0}
\newcommand{\Mu}{M_u}
\newcommand{\vuModel}{Qwen3-VL-8B}
\newcommand{\iv}[2]{$[#1,\,#2]$}

\newcommand{\takeaway}[1]{%
\vspace{3pt}
\noindent\fbox{%
\parbox{\dimexpr\columnwidth-2\fboxsep-2\fboxrule}{%
\small\textbf{Takeaway.} #1
}}
\vspace{4pt}
}

\title{Suppression Is Not Forgetting: Residual Recoverability in Visual Concept Unlearning for VLMs}

\name{Zhangyun Tan$^{\dagger}$, Zeliang Zhang$^{\dagger}$, Jiani Liu,  Susan Liang, Yolo Y. Tang, Lisha Chen, Chenliang Xu\thanks{$^{\dagger}$Equal contribution. Code is available at \url{https://github.com/zhangyun04/ULBench}.}}
\address{University of Rochester}

\begin{document}
\ninept
\maketitle

\begin{abstract}
Vision-language models (VLMs) may need to forget visual concepts after deployment because of privacy, copyright, licensing, safety, or policy changes. Conventional machine unlearning modifies model parameters, which may be costly or inaccessible for API-only models. Prompt-based suppression offers a training-free alternative, but does it make a concept inaccessible or merely change the model's answer?

We investigate this question in off-the-shelf VLMs. Our visually grounded, multi-probe evaluation first verifies that a model recognizes each concept from the image, then tests its recoverability through multiple-choice, short-answer, and indirect queries. Across objects, scenes, and identities, prompt suppression reduces short-answer recall for some concepts while leaving multiple-choice and indirect performance largely unchanged. Explicitly listing the concepts to suppress often increases short-answer recall, suggesting that the list itself cues the answer. Beyond prompting, decoding constraints, representation editing, and parameter updates can suppress particular responses while the same concept remains detectable through other queries. A model may stop naming a visual concept yet still identify or use it when queried differently.

\end{abstract}


\section{Introduction}
\label{sec:intro}

VLMs trained on web-scale data recognize people, logos, objects, and scenes that a deployment may later be required to stop recognizing due to privacy, copyright, safety, or policy concerns.
Machine unlearning studies how to remove such knowledge from a trained model~\cite{bourtoule2021machine,nguyen2025survey}.
Existing multimodal unlearning benchmarks~\cite{liu2025mllmu,dontsov2024clear,ma2024fiubench} often introduce synthetic knowledge through fine-tuning and then evaluate whether it can be removed.
A common deployment setting is different: the concept is already present in the model, and the central question is whether an intervention makes that concept inaccessible, or whether it remains recoverable in other forms despite a change in the model's response.

This distinction is difficult to establish from output accuracy alone.
A meaningful test of visual concept unlearning must satisfy several requirements.
First, the target concept must be demonstrably grounded in the image rather than recoverable from language priors.
Second, forgetting should persist across different ways of accessing the same concept, rather than only one answer format.
Third, failures of instruction following, refusal, or decoding should not be conflated with loss of knowledge.
Fourth, evaluation should reflect semantic access rather than superficial string mismatch.
Finally, an unlearning effect should be robust to training stochasticity rather than depend on a favorable optimization run.
Together, these requirements shift the question from whether a model produces a particular answer to whether the concept remains behaviorally recoverable.

We develop an evaluation framework around this view and study interventions
spanning prompting, decoding, representation editing, and parameter updates.
Our results reveal three consistent patterns.
First, \textbf{suppression leaves substantial residual recoverability}: prompt-based interventions can reduce explicit naming while the same concept remains recognizable through alternative queries.
Second, \textbf{stronger suppression need not imply stronger forgetting}: explicitly presenting the concepts to be suppressed can instead act as a retrieval cue and increase their disclosure.
Third, \textbf{the gap between suppression and forgetting persists beyond prompting}: decoding constraints, representation editing, and weight updates can each reduce access through particular probes without reliably eliminating the concept across probes, and apparent success can be sensitive to training randomness.

These findings suggest that visual concept unlearning should be evaluated by
whether the target remains recoverable across complementary probes, rather than by reduced disclosure under a single prompt, scorer, or training run.

\section{Related work.}
Text-only benchmarks such as TOFU, MUSE, RWKU, and WMDP~\cite{maini2024tofu,shi2024muse,cao2024rwku,li2024wmdp} established multi-probe evaluation; RWKU targets knowledge already present in the model.
Prompt- and input-based alternatives avoid direct weight updates, including in-context unlearning~\cite{pawelczyk2023context}, guardrail prompting and filtering~\cite{thaker2024guardrail}, and learned soft prompts~\cite{bhaila2025soft}.
Takashiro et al.~\cite{takashiro2025answer} show that an LLM can ``pretend to forget,'' retaining the target answer internally while suppressing it at the output, while black-box evaluations show that apparently forgotten knowledge can often be recovered by rephrasing or few-shot prompting~\cite{doshi2024does}.
MLLMU-Bench, CLEAR, and FIUBench~\cite{liu2025mllmu,dontsov2024clear,ma2024fiubench} inject fictitious identities by fine-tuning; PPU-Bench~\cite{guang2026ppubench} evaluates pre-existing knowledge about public figures, while AUVIC~\cite{chen2026auvic} directly studies visual concept unlearning in MLLMs.
Multimodal unlearning methods span gradient-based updates~\cite{huo2025mmunlearner,li2024siu} and training-free interventions on activations or MLP vectors~\cite{ding2025mllmeraser,kim2026kvw}; Lin et al.~\cite{lin2026robustness} show that supposedly forgotten multimodal knowledge can be reactivated by contextual prompting or downstream retraining.
We instead ask whether visual concepts remain behaviorally recoverable after
unlearning, using visual-grounding controls and varying the query format,
scoring rule, and optimization randomness.

\section{Measuring Visual Concept Forgetting}
\label{sec:protocol}

We design a benchmark to measure \emph{residual recoverability} after visual
concept unlearning: if a concept disappears from one response, can it still be
recovered through another?
To answer this, we probe the same concept through three complementary abilities:
\emph{recognition}, \emph{naming}, and \emph{use}.

\subsection{How to probe visual concept knowledge?}

We study visual concepts from three broad categories---objects, scenes, and
identities---using images from COCO~\cite{lin2014microsoft},
AID~\cite{xia2016aid}, and a celebrity face dataset~\cite{celebrity_faces}.
For every test image, we construct three probes of the same underlying concept:
\begin{itemize}
    \item A multiple-choice (MCQ) probe asks whether the model can
    \emph{recognize} the concept among alternatives.
    \item A short-answer (SA) probe asks whether the model can
    \emph{name} it directly.
    \item An indirect probe asks a related question whose answer depends on the
    concept without requiring its name.
\end{itemize}

These three probes are intentionally complementary.
A model may refuse to say ``zebra'' while still selecting zebra from four
choices, or may avoid naming a person while still answering a question about
that person's nationality.
In such cases, the output has changed, but the concept remains recoverable through another query.

We also construct no-image, option-only, and shuffled-image controls.
They test whether the answer truly comes from the image rather than from
language priors, answer choices, or dataset regularities.
All probes from the same source image are paired, so the untreated and
intervened models are compared on exactly the same examples.
Refusals, invalid outputs, and operational errors are recorded separately
rather than counted as forgetting.

\noindent\textbf{Verifying visual grounding.}
Not every candidate concept provides meaningful evidence of forgetting.
If a model cannot recognize a concept before the intervention, or can answer the
question without using the image, a later accuracy drop says little about
unlearning.
We therefore evaluate only concepts that are \emph{visually grounded} in the
untreated model $\Mzero$.

\begin{figure*}[t]
\centering
\includegraphics[width=0.95\textwidth]{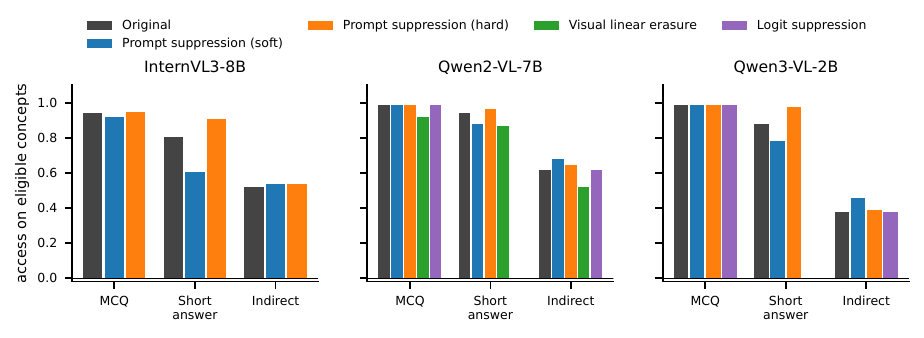}
\caption{
Access to the same visual concepts through multiple probes before and after
prompt-, decoding-, and representation-level interventions.
Bars show mean access over eligible concepts and dots show individual concepts.
Indirect access is prior-adjusted.
}
\label{fig:leakage}
\end{figure*}

A concept is included when the model performs reliably on the normal-image
probes, clearly outperforms the no-image, option-only, and shuffled-image
controls, and gives consistent answers across paraphrased questions.
In other words, before testing whether a concept has been forgotten, we first
establish that the model actually knew it from the image.

Concretely, normal-image access must exceed
$\tau_{\mathrm{acc}}$ (0.60 for MCQ and 0.50 for SA), and its gap over the
strongest control must exceed $\tau_{\mathrm{vis}}=0.20$.
All thresholds are fixed before evaluating any intervention.

\subsection{How to measure forgetting?}
\label{sec:metrics}

Once a concept is verified to be visually grounded, we compare the model
before and after applying a candidate forgetting method on the same images.
For a concept $c$ and probe type $g$, let $A_g^0(c)$ and $A_g^u(c)$ denote the
fraction of paired examples answered correctly before and after the method is
applied.
We call this quantity the \emph{access rate}: it measures how often the model
can still recover the concept through a particular probe.
We report the access drop
\[
\Delta_g(c)=A_g^0(c)-A_g^u(c),
\]
where a positive value means that the concept has become harder to recover
under that probe. 

A large drop on one probe, however, does not by itself imply forgetting.
The same concept may disappear from one type of answer while remaining
available through another.
For example, a model may stop producing ``zebra'' in a short answer but still
select zebra correctly from multiple choices.
Likewise, it may avoid naming a person while still answering questions that
depend on knowing who that person is.
We therefore look at the pattern across MCQ, short-answer, and indirect probes,
rather than treating any single score as evidence of forgetting.

Two challenges remain before these measurements can be interpreted as evidence
of forgetting.
First, an indirect question may sometimes be answered without looking at the
image.
To separate visual information from language-only cues, we evaluate the same
question with the image removed and subtract that performance from the
normal-image result.
We refer to the resulting quantity as \emph{prior-adjusted indirect access}.

Second, generated answers are not always well captured by exact string
matching.
For example, answering only a person's first name may fail an exact-match
scorer even though the response still reveals part of the identity.
We therefore report both a strict exact-match scorer and a rule-based semantic
scorer.
The semantic scorer matches each response against a concept-specific catalog
containing the target name, aliases, translations, and related entities, and
labels it as complete identification, partial identification, absence, or
ambiguous.

Let $L(M)$ count only complete identifications and let $U(M)$ additionally count
partial and ambiguous responses as possible disclosure.
For an intervention $u$, we report the semantic disclosure drop as
\[
\left[
L(\Mzero)-U(\Mu),\;
U(\Mzero)-L(\Mu)
\right].
\]
The lower bound is the conservative estimate of forgetting, while the upper
bound is the permissive estimate. To compare response formats directly, we report the \emph{cross-probe gap}
$\Delta_{\mathrm{SA}}-\Delta_{\mathrm{MCQ}}$, using the semantic lower bound
for $\Delta_{\mathrm{SA}}$, together with a 95\% bootstrap confidence interval
over concepts and images.

Together, these steps reduce false evidence of forgetting caused by language
priors or scoring artifacts.

\subsection{Interventions under evaluation}
\label{sec:interventions}

We evaluate the framework across interventions ranging from
prompt-based suppression to representation editing and parameter updates,
allowing us to examine whether residual recoverability persists as the
intervention acts more directly on the model.

\noindent\textbf{Prompt and decoding.}
Following guardrail and in-context unlearning baselines
~\cite{thaker2024guardrail,pawelczyk2023context}, we evaluate two generic
suppression prompts carrying the target concept list.
The \emph{soft} prompt asks the model not to identify or reveal any listed
concept and to choose a different answer when necessary, whereas the
\emph{hard} prompt states that the listed classes have been unlearned and
explicitly forbids selecting them.
We evaluate both prompts on Qwen3-VL-2B~\cite{bai2025qwen3},
InternVL3-8B~\cite{zhu2025internvl3}, and
Qwen2-VL-7B~\cite{wang2024qwen2}.
We also apply \emph{logit suppression} on Qwen2-VL-7B and Qwen3-VL-2B to
directly block generation of the target names.

\noindent\textbf{Representation editing.}
We apply \emph{visual linear erasure}, estimating target-concept directions
from training images by differences in means and projecting visual
tokens onto their orthogonal complement before language generation,
inspired by mean-projection concept erasure~\cite{haghighatkhah2022better}.

\noindent\textbf{Parameter-update methods.}
We evaluate gradient ascent~\cite{jang2023knowledge}, gradient difference and
KL-regularized ascent~\cite{maini2024tofu,liu2025mllmu},
NPO~\cite{zhang2024npo}, and MMUnlearner~\cite{huo2025mmunlearner}.
Methods are selected on validation data to balance forgetting of the target
concepts and preservation of matched retain concepts.
For GA, NPO, and MMUnlearner, we additionally test whether the selected
configuration remains effective across independent training runs
(Sec.~\ref{sec:robutenss}).

\begin{table*}[t]
\centering
\caption{
Parameter-update unlearning on \vuModel.
Target denotes concepts to be unlearned and Retain denotes matched non-target
concepts.
Positive values indicate reduced access after unlearning.
$\Delta$SA is the semantic disclosure interval and $\Delta$MCQ is the change
in generative multiple-choice access.
The cross-probe gap is the lower-bound $\Delta$SA minus $\Delta$MCQ; a positive
95\% confidence interval indicates a larger reduction in short-answer disclosure
than in multiple-choice recognition.
}
\label{tab:weight}
\scriptsize
\setlength{\tabcolsep}{4pt}
\begin{tabular}{@{}lccccc@{}}
\toprule
Method &
Target $\Delta$SA [LB,UB] &
Target $\Delta$MCQ &
Cross-probe gap, 95\% CI &
Retain $\Delta$SA $[LB,UB]$ &
Retain $\Delta$MCQ \\
\midrule
GA &
\iv{0.040}{0.201} &
$0.000$ &
$0.040$ \iv{-0.025}{0.117} &
\iv{-0.008}{0.033} &
$0.008$ \\
GD &
\iv{-0.022}{0.024} &
$0.000$ &
$-0.023$ \iv{-0.069}{0.022} &
\iv{-0.025}{0.008} &
$0.000$ \\
GA+KL &
\iv{-0.019}{0.034} &
$0.000$ &
$-0.020$ \iv{-0.083}{0.047} &
\iv{-0.017}{0.017} &
$0.008$ \\
NPO &
\iv{0.040}{0.215} &
$0.000$ &
$0.040$ \iv{-0.025}{0.111} &
\iv{-0.008}{0.033} &
$0.008$ \\
MMUnlearner &
\iv{0.135}{0.173} &
$0.027$ &
$0.108$ \iv{0.018}{0.217} &
\iv{-0.033}{-0.017} &
$0.033$ \\
\bottomrule
\end{tabular}
\end{table*}

\section{Results}
\label{sec:results}

\subsection{Settings}

\noindent\textbf{Visual grounding.}
We evaluate 15 visual concepts spanning objects, scenes, and identities.
After applying the visual-grounding criteria of Sec.~\ref{sec:protocol},
13 concepts are eligible for Qwen3-VL-2B, 10 for InternVL3-8B, and 14 for
Qwen2-VL-7B.
On the eligible concepts, normal-image MCQ performance is near ceiling while
the strongest controls remain substantially lower, confirming that the
evaluation depends on the visual input rather than language or option priors.
Short-answer controls are zero across all three models.
A human audit of the exact short-answer scorer gives 0.957 agreement, with the
remaining disagreements mainly arising from descriptive scene responses.

\noindent\textbf{Evaluation setup.}
We evaluate prompt suppression on all three models, and additionally study
logit suppression and visual linear erasure on Qwen2-VL-7B.
For these interventions, an MCQ is correct when the target option receives the
highest log-probability, allowing us to measure recognition even when generation
is constrained.
We further evaluate five parameter-update methods on \vuModel; here MCQ is
generative, so the model must produce its selected answer.
Unless stated otherwise, all comparisons use the paired examples defined in
Sec.~\ref{sec:metrics}.

\subsection{Suppression does not generalize across probes}

\textbf{Prompt suppression primarily changes what the model says, not what it
can still identify.}
Under the softer suppression prompt, short-answer access decreases on all three
models, whereas MCQ access changes by at most 0.022.
Indirect access also remains close to the untreated level.
Thus, a model can become less likely to \emph{name} a concept while continuing
to recognize and use it when queried differently.

\textbf{Stronger suppression can instead increase disclosure.}
When the prompt explicitly lists the concepts to suppress, short-answer recall
increases on 8/13 eligible concepts for Qwen3-VL-2B, 6/14 for Qwen2-VL-7B,
and 8/10 for InternVL3-8B.
Placing the forbidden concepts in context can therefore act as a retrieval cue
rather than a block.

\textbf{Directly blocking generation still does not remove recognition.}
Logit suppression drives short-answer access on Qwen2-VL-7B from 0.94 to 0.00,
yet leaves MCQ and indirect access unchanged.
The target name can no longer be generated, but the model still recognizes and
uses the same concept under other probes.
Visual linear erasure has a broader effect, lowering all three probes
simultaneously, while MMMU accuracy remains essentially unchanged
(0.509 versus 0.510 untreated).

\takeaway{
Suppressing one way of expressing a concept does not make the concept
unavailable through other queries.
}

\subsection{Parameter updates remain probe- and concept-dependent}

\textbf{The cross-probe mismatch persists even after updating model parameters.}
As shown in Table~\ref{tab:weight}, generative MCQ access changes little under
all five methods, with a maximum drop of 0.027, even when short-answer
disclosure decreases.
Thus, changing the model weights does not necessarily remove its ability to
identify the target when the same concept is queried in a different form.

The extent of this mismatch varies across methods.
Gradient difference and KL-regularized ascent show no reliable reduction in
semantic disclosure.
Gradient ascent and NPO reduce short-answer disclosure for some concepts, but
their cross-probe confidence intervals include zero.
MMUnlearner is the only method whose cross-probe gap has a 95\% confidence
interval entirely above zero, indicating that its short-answer reduction is
reliably larger than its MCQ reduction.

\textbf{The effect is also strongly concept-dependent.}
GA and NPO show a positive lower-bound disclosure reduction on only four of the
fourteen eligible concepts, and MMUnlearner on seven; all COCO object concepts
remain nearly unchanged under every method.
Parameter updates therefore produce measurable forgetting for some concepts,
but not a uniform loss of access across probes or concepts.

A paired example makes this distinction concrete.
For the same Kate Winslet image, the untreated model answers ``Kate Winslet,''
whereas MMUnlearner answers ``Elizabeth McGovern'' in free form but still
selects Kate Winslet from four candidates.
This advantage disappears when the image is removed, showing that the
intervened model retains visually grounded recognition even when its free-form
answer changes.

\takeaway{
Parameter updates can reduce disclosure, but the target concept can remain
recoverable through other probes and for many individual concepts.
}

\begin{table}[t]
\centering
\caption{
The same generated outputs yield different estimates of the short-answer
disclosure drop under different scoring rules.
Exact uses alias matching; Lexical additionally counts partial name matches as
disclosure; Semantic reports the lower (LB) and upper (UB) bounds defined in
Sec.~\ref{sec:metrics}.
}
\label{tab:ladder}
\setlength{\tabcolsep}{8pt}
\begin{tabular}{@{}lrrrr@{}}
\toprule
& & & \multicolumn{2}{c}{Semantic} \\
\cmidrule(lr){4-5}
Method & Exact & Lexical & LB & UB \\
\midrule
GA      & 0.192 & 0.089 &  0.040 & 0.201 \\
GD      & 0.031 & 0.031 & -0.022 & 0.024 \\
GA+KL   & 0.090 & 0.067 & -0.019 & 0.034 \\
NPO     & 0.210 & 0.084 &  0.040 & 0.215 \\
MMUnl.  & 0.134 & 0.143 &  0.135 & 0.173 \\
\bottomrule
\end{tabular}
\end{table}

\begin{figure}[t]
\centering
\includegraphics[width=0.92\columnwidth]{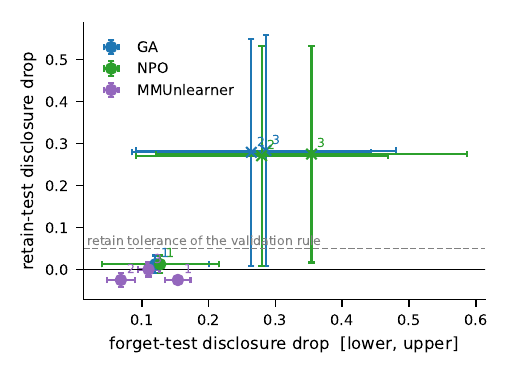}
\caption{
Robustness of selected parameter-update configurations across independent
training runs.
The x- and y-axes show test-set semantic disclosure drops on target and retain
concepts, with error bars spanning the corresponding semantic lower and upper
bounds.
The dashed line marks the SA retain tolerance used during model selection.
Circles denote runs satisfying the full preservation criterion, crosses denote
failures, and numbers indicate random seeds.
}
\label{fig:seeds}
\end{figure}

\subsection{Robustness of Unlearning Evaluation}
\label{sec:robutenss}
We next examine whether an unlearning conclusion remains stable under two
sources of variation: the scoring rule used to interpret generated outputs and
the randomness of optimization.

\noindent\textbf{Sensitivity to scoring.}
Table~\ref{tab:ladder} shows that the same generated outputs can lead to
substantially different estimates of forgetting.
Under exact matching, GA and NPO show short-answer disclosure drops of 0.192
and 0.210.
After accounting for partial names and other semantically revealing responses,
their conservative semantic lower bounds fall to 0.040.
Thus, part of the apparent forgetting under exact matching reflects changes in
surface form rather than disappearance of the target information.
MMUnlearner is less sensitive to the scoring rule, with a semantic lower bound
of 0.135 compared with an exact-match drop of 0.134.

\noindent\textbf{Sensitivity to optimization randomness.}
Fig.~\ref{fig:seeds} tests whether a configuration selected on one run remains
valid when retrained independently.
GA and NPO satisfy the preservation criterion on the selection run but fail it
on both independent reruns.
These failures coincide with severe response truncation and large losses on
retain concepts, indicating generation collapse rather than selective
forgetting.
MMUnlearner satisfies the preservation criterion across all runs, although the
magnitude of its forgetting effect still varies substantially.

\takeaway{
A credible unlearning result should be robust to both the scoring rule and
ordinary optimization randomness.
}

\section{Conclusion}
\label{sec:conclusion}

We study visual concept unlearning through \emph{residual recoverability}:
whether a concept that disappears from one response can still be recovered
through another.
Across prompt, decoding, representation, and parameter-update interventions,
reduced disclosure often coexists with preserved recognition or indirect use.
The apparent degree of forgetting can also depend on scoring and optimization
randomness.
These results show that suppressing one response is not sufficient evidence of
forgetting, motivating visually grounded, complementary probes and robustness
to evaluation and training variations.

\bibliographystyle{IEEEbib}
\bibliography{refs}

@inproceedings{chen2026auvic,
  title={AUVIC: Adversarial unlearning of visual concepts for multi-modal large language models},
  author={Chen, Haokun and Li, Jianing and Zhang, Yao and Bi, Jinhe and Xia, Yan and Gu, Jindong and Tresp, Volker},
  booktitle={Proceedings of the AAAI Conference on Artificial Intelligence},
  volume={40},
  number={36},
  pages={30218--30225},
  year={2026}
}

@article{doshi2024does,
  title={Does unlearning truly unlearn? a black box evaluation of llm unlearning methods},
  author={Doshi, Jai and Stickland, Asa Cooper},
  journal={arXiv preprint arXiv:2411.12103},
  year={2024}
}

@inproceedings{takashiro2025answer,
  title={Answer when needed, forget when not: Language models pretend to forget via in-context knowledge unlearning},
  author={Takashiro, Shota and Kojima, Takeshi and Gambardella, Andrew and Cao, Qi and Iwasawa, Yusuke and Matsuo, Yutaka},
  booktitle={Findings of the Association for Computational Linguistics: ACL 2025},
  pages={24872--24885},
  year={2025}
}

@inproceedings{bhaila2025soft,
  title={Soft prompting for unlearning in large language models},
  author={Bhaila, Karuna and Van, Minh-Hao and Wu, Xintao},
  booktitle={Proceedings of the 2025 Conference of the Nations of the Americas Chapter of the Association for Computational Linguistics: Human Language Technologies (Volume 1: Long Papers)},
  pages={4046--4056},
  year={2025}
}

@article{wang2024qwen2,
  title={Qwen2-vl: Enhancing vision-language model's perception of the world at any resolution},
  author={Wang, Peng and Bai, Shuai and Tan, Sinan and Wang, Shijie and Fan, Zhihao and Bai, Jinze and Chen, Keqin and Liu, Xuejing and Wang, Jialin and Ge, Wenbin and others},
  journal={arXiv preprint arXiv:2409.12191},
  year={2024}
}

@inproceedings{bourtoule2021machine,
  title={Machine unlearning},
  author={Bourtoule, Lucas and Chandrasekaran, Varun and Choquette-Choo, Christopher A and others},
  booktitle={2021 IEEE symposium on security and privacy (SP)},
  pages={141--159},
  year={2021},
  organization={IEEE}
}

@article{nguyen2025survey,
  title={A survey of machine unlearning},
  author={Nguyen, Thanh Tam and Huynh, Thanh Trung and Ren, Zhao and others},
  journal={ACM Transactions on Intelligent Systems and Technology},
  volume={16},
  number={5},
  pages={1--46},
  year={2025},
  publisher={ACM New York, NY}
}

@inproceedings{jang2023knowledge,
  title={Knowledge unlearning for mitigating privacy risks in language models},
  author={Jang, Joel and Yoon, Dongkeun and Yang, Sohee and others},
  booktitle={Proceedings of the 61st Annual Meeting of the Association for Computational Linguistics (Volume 1: Long Papers)},
  pages={14389--14408},
  year={2023}
}

@article{maini2024tofu,
  title={Tofu: A task of fictitious unlearning for llms},
  author={Maini, Pratyush and Feng, Zhili and Schwarzschild, Avi and Lipton, Zachary C and Kolter, J Zico},
  journal={arXiv preprint arXiv:2401.06121},
  year={2024}
}

@article{shi2024muse,
  title={Muse: Machine unlearning six-way evaluation for language models},
  author={Shi, Weijia and Lee, Jaechan and Huang, Yangsibo and others},
  journal={arXiv preprint arXiv:2407.06460},
  year={2024}
}

@article{cao2024rwku,
  title={Rwku: Benchmarking real-world knowledge unlearning for large language models},
  author={Cao, Pengfei and Wang, Chenhao and He, Zhitao and Yuan, Hongbang and Li, Jiachun and Chen, Yubo and Liu, Kang and Zhao, Jun and others},
  journal={Advances in Neural Information Processing Systems},
  volume={37},
  pages={98213--98263},
  year={2024}
}

@article{li2024wmdp,
  title={The wmdp benchmark: Measuring and reducing malicious use with unlearning},
  author={Li, Nathaniel and Pan, Alexander and Gopal, Anjali and Yue, Summer and Berrios, Daniel and Gatti, Alice and Li, Justin D and Dombrowski, Ann-Kathrin and Goel, Shashwat and Phan, Long and others},
  journal={arXiv preprint arXiv:2403.03218},
  year={2024}
}

@article{pawelczyk2023context,
  title={In-context unlearning: Language models as few shot unlearners},
  author={Pawelczyk, Martin and Neel, Seth and Lakkaraju, Himabindu},
  journal={arXiv preprint arXiv:2310.07579},
  year={2023}
}

@article{thaker2024guardrail,
  title={Guardrail baselines for unlearning in llms},
  author={Thaker, Pratiksha and Maurya, Yash and Hu, Shengyuan and Wu, Zhiwei Steven and Smith, Virginia},
  journal={arXiv preprint arXiv:2403.03329},
  year={2024}
}

@inproceedings{lin2014microsoft,
  title={Microsoft coco: Common objects in context},
  author={Lin, Tsung-Yi and Maire, Michael and Belongie, Serge and Hays, James and Perona, Pietro and Ramanan, Deva and Doll{\'a}r, Piotr and Zitnick, C Lawrence},
  booktitle={European conference on computer vision},
  pages={740--755},
  year={2014},
  organization={Springer}
}

@article{xia2016aid,
  title={{AID}: A benchmark dataset for performance evaluation of aerial scene classification},
  author={Xia, G and Hu, J and Hu, F and others},
  journal={arXiv preprint arXiv:1608.05167},
  year={2016}
}

@misc{celebrity_faces,
    title={Celebrity Face Image Dataset},
    author={Vishesh, Verma},
    year={2022},
    howpublished={\url{https://www.kaggle.com/datasets/vishesh1412/celebrity-face-image-dataset}}
  }

@article{zhu2025internvl3,
  title={Internvl3: Exploring advanced training and test-time recipes for open-source multimodal models},
  author={Zhu, Jinguo and Wang, Weiyun and Chen, Zhe and Liu, Zhaoyang and Ye, Shenglong and Gu, Lixin and Tian, Hao and Duan, Yuchen and Su, Weijie and Shao, Jie and others},
  journal={arXiv preprint arXiv:2504.10479},
  year={2025}
}

@article{bai2025qwen3,
  title={Qwen3-vl technical report},
  author={Bai, Shuai and Cai, Yuxuan and Chen, Ruizhe and Chen, Keqin and Chen, Xionghui and Cheng, Zesen and Deng, Lianghao and Ding, Wei and Gao, Chang and Ge, Chunjiang and others},
  journal={arXiv preprint arXiv:2511.21631},
  year={2025}
}

@inproceedings{liu2025mllmu,
  title={Protecting privacy in multimodal large language models with {MLLMU-Bench}},
  author={Liu, Zheyuan and Dou, Guangyao and Jia, Mengzhao and others},
  booktitle={Proceedings of NAACL},
  year={2025}
}

@article{dontsov2024clear,
  title={{CLEAR}: Character unlearning in textual and visual modalities},
  author={Dontsov, Alexey and Korzh, Dmitrii and Zhavoronkin, Alexey and others},
  journal={arXiv preprint arXiv:2410.18057},
  year={2024}
}

@article{ma2024fiubench,
  title={Benchmarking vision language model unlearning via fictitious facial identity dataset},
  author={Ma, Yingzi and Wang, Yulong and Zhang, Yibo and Wu, Ruoxi and Xu, Feiran and Ge, Zhiyuan and Wang, Xiaoyu and Tan, Yao and Li, Bo and Zhao, Ping and others},
  journal={arXiv preprint arXiv:2411.03554},
  year={2024}
}

@inproceedings{huo2025mmunlearner,
  title={{MMUnlearner}: Reformulating multimodal machine unlearning in the era of multimodal large language models},
  author={Huo, Jiahao and Yan, Yibo and Hu, Xu and others},
  booktitle={Findings of ACL},
  year={2025}
}

@article{li2024siu,
  title={Single image unlearning: Efficient machine unlearning in multimodal large language models},
  author={Li, Jiaqi and Wei, Qianshan and Zhang, Chuanyi and others},
  journal={arXiv preprint arXiv:2405.12523},
  year={2024}
}

@article{zhang2024npo,
  title={Negative preference optimization: From catastrophic collapse to effective unlearning},
  author={Zhang, Ruiqi and Lin, Licong and Bai, Yu and Mei, Song},
  journal={arXiv preprint arXiv:2404.05868},
  year={2024}
}

@article{guang2026ppubench,
  title={{PPU-Bench}: Real world benchmark for personalized partial unlearning in vision language models},
  author={Guang, Jiahui and Zhan, Zexun and Xu, Zhenlin and others},
  journal={arXiv preprint arXiv:2605.08800},
  year={2026}
}

@article{ding2025mllmeraser,
  title={{MLLMEraser}: Achieving test-time unlearning in multimodal large language models through activation steering},
  author={Ding, Chenlu and Wu, Jiancan and Sheng, Leheng and others},
  journal={arXiv preprint arXiv:2510.04217},
  year={2025}
}

@article{kim2026kvw,
  title={Knowledge vector weakening: Efficient training-free unlearning for large vision-language models},
  author={Kim, Yejin and Hwang, Dongjun and Cha, Sungmin and Choe, Junsuk},
  journal={arXiv preprint arXiv:2601.21794},
  year={2026}
}

@article{lin2026robustness,
  title={On the robustness of machine unlearning for vision-language models},
  author={Lin, Yujie and Jia, Kaidi and Ma, Jiayao and Yang, Chengyi and Su, Jinsong},
  journal={arXiv preprint arXiv:2605.26992},
  year={2026}
}

@inproceedings{haghighatkhah2022better,
  title     = {Better Hit the Nail on the Head than Beat around the Bush: Removing Protected Attributes with a Single Projection},
  author    = {Haghighatkhah, Pantea and Fokkens, Antske and Sommerauer, Pia and Speckmann, Bettina and Verbeek, Kevin},
  booktitle = {Proceedings of the 2022 Conference on Empirical Methods in Natural Language Processing},
  pages     = {8395--8416},
  year      = {2022},
  doi       = {10.18653/v1/2022.emnlp-main.575}
}

\end{document}